\documentclass[letterpaper]{article} 
\usepackage{aaai2026}  
\usepackage{times}  
\usepackage{helvet}  
\usepackage{courier}  
\usepackage[hyphens]{url}  
\usepackage{graphicx} 
\usepackage{natbib}  
\usepackage{caption} 
\usepackage{subfigure}
\usepackage{multicol}
\usepackage{booktabs} 
\usepackage{amssymb}
\usepackage{subcaption}  
\usepackage{adjustbox}
\usepackage{float}
\usepackage{pifont}
\usepackage{graphicx}
\usepackage{amsmath}
\usepackage{amssymb}
\usepackage{booktabs}
\usepackage{xcolor}
\usepackage{multirow}
\usepackage{pifont}
\usepackage{algorithm}
\usepackage{algorithmic}

\usepackage{newfloat}
\usepackage{listings}
\DeclareCaptionStyle{ruled}{labelfont=normalfont,labelsep=colon,strut=off} 
\floatstyle{ruled}
\newfloat{listing}{tb}{lst}{}
\floatname{listing}{Listing}
\nocopyright

\title{Can Large Pretrained Depth Estimation Models Help With Image Dehazing?}
\author{
    Hongfei Zhang,
    Kun Zhou,
    Ruizheng Wu,
    Jiangbo Lu\thanks{Jiangbo Lu is the corresponding author.}
}
\affiliations{
    SmartMore Corporation \\

    hongfei@u.nus.edu,
    zhoukun303808@gmail.com,
    ruizheng.wu@smartmore.com,
    jiangbo.lu@gmail.com

}

\usepackage{bibentry}

\begin{document}

\maketitle

\begin{abstract}
    Image dehazing remains a challenging problem due to the spatially varying nature of haze in real-world scenes. While existing methods have demonstrated the promise of large-scale pretrained models for image dehazing, their architecture-specific designs hinder adaptability across diverse scenarios with different accuracy and efficiency requirements. In this work, we systematically investigate the generalization capability of pretrained depth representations—learned from millions of diverse images—for image dehazing. Our empirical analysis reveals that the learned deep depth features maintain remarkable consistency across varying haze levels. Building on this insight, we propose a plug-and-play RGB-D fusion module that seamlessly integrates with diverse dehazing architectures. Extensive experiments across multiple benchmarks validate both the effectiveness and broad applicability of our approach.

\end{abstract}


\section{Introduction}

Image dehazing is a fundamental task in computer vision, aiming to recover clear and accurate visual information from images degraded by atmospheric particles such as haze, fog, or smoke.
It is essential in many real-world applications, including autonomous driving, surveillance, and image enhancement, where visibility plays a critical role. Traditional dehazing methods typically rely on statistical priors that assume specific physical properties of hazy images~\cite{schaul2009color,kim2011single,fattal2008single,he2010single}. For example, \citeauthor{fattal2008single} proposed a method based on the local uncorrelation between transmission and surface shading, using an enhanced model to estimate scene transmission. \citeauthor{he2010single} introduced the dark channel prior, which assumes that at least one color channel has very low intensity. However, these priors are often constrained by various conditions, limiting their reliability in complex real-world scenarios.

\begin{figure}[h]
    \centering

        \includegraphics[width=\linewidth]{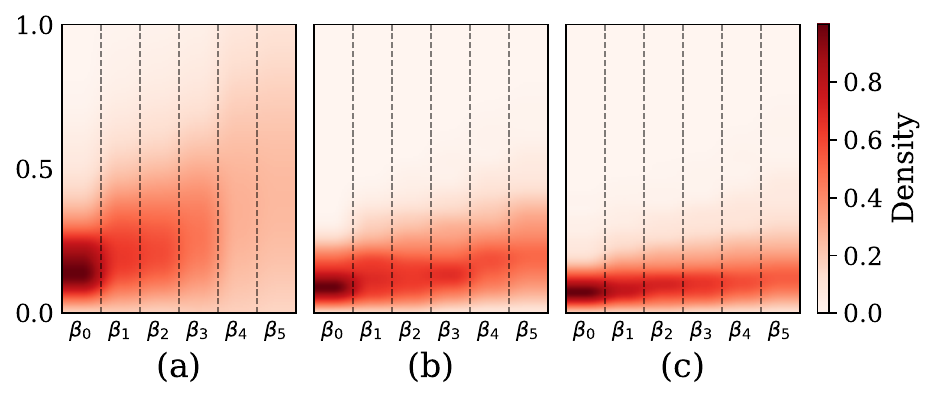}

        \caption{
            Heatmaps illustrating the normalized distributions of distances between deep features of hazy images and their corresponding ground truth across six haze levels, ranging from light to heavy haze. Here, $\beta$ denotes the atmospheric scattering coefficient~\cite{mccartney1976optics}, varying from 0.04 to 0.20. (a) Baseline~\cite{fang2025guided} without depth; (b) Baseline enhanced by~\cite{ye2025prompthaze}; (c) Baseline enhanced by our method. Results are computed over 10k images from the RESIDE dataset~\cite{li2019benchmarking}. It shows that incorporating depth consistently improves feature alignment across varying haze levels, with our method achieving the most compact and stable feature distributions.
        }
        
    \label{fig:tsne-vis}
\end{figure}

Recent advances in deep learning, especially CNN- and Transformer-based models, have significantly improved image dehazing~\cite{zhu2015fast,qin2020ffa,liu2019griddehazenet,tu2022maxim,wu2021contrastive,guo2022image,chen2021psd,zheng2023curricular}. These methods leverage large datasets to learn haze-to-clear mappings effectively. However, purely data-driven approaches often struggle to capture geometry information, due to their reliance on training data and lack of physical priors~\cite{wu2021contrastive,qin2020ffa,yang2022depth,zhang2024depth,fang2025guided}. To address this, previous works try to integrate depth cues into dehazing , either by jointly predicting depth and clear images~\cite{yang2022depth} or by employing dual-task strategies~\cite{zhang2024depth}. Nevertheless, such approaches rely on depth estimators trained on limited, scene-specific datasets, which constrains their generalization capability. In contrast, large-scale monocular depth estimation models~\cite{yang2024depth,yang2024depthv2,he2025distill}, trained on diverse scenes, produce structurally consistent depth features, offering a reliable complementary signal for improving dehazing quality.

\begin{figure*}[h]
    \centering
    \includegraphics[width=\linewidth]{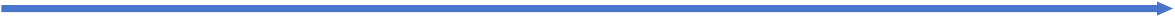}
    \includegraphics[width=\linewidth]{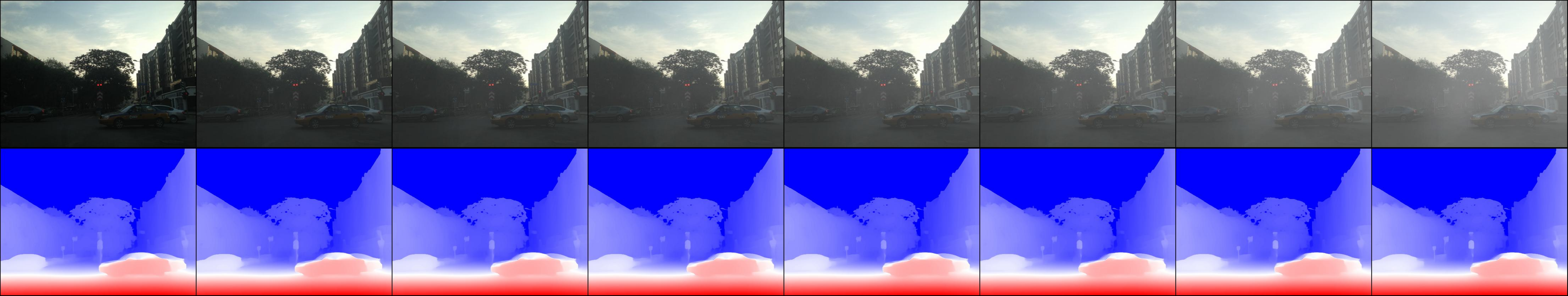}
    \includegraphics[width=\linewidth]{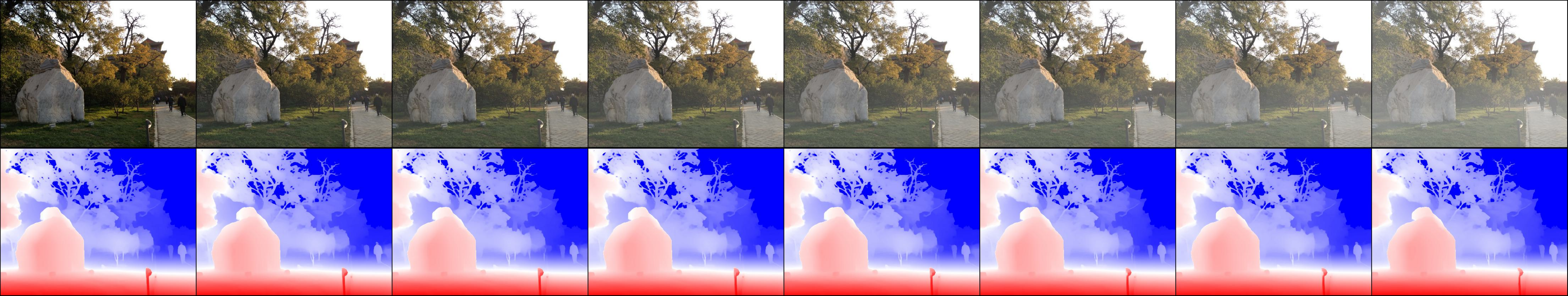}
    \caption{
        Depth prediction results from a large pretrained depth estimation model under varying haze levels for two challenging scene.
        Despite significant changes in haze density, the depth maps remain remarkably consistent, demonstrating its consistency to visibility degradation. 
        Additional examples are provided in the appendix.
    }
    \label{fig:intro-multi-haze-compare}
\end{figure*}

Recent studies~\cite{chen2024teaching, ye2025prompthaze} have explored the integration of large-scale monocular depth estimation models into image dehazing. However, these approaches face several limitations. Primarily, they overlook a thorough analysis or understanding of how depth features contribute to the dehazing process. Additionally, their methods are tailored to specific network architectures, reducing generalizability and hindering broader applicability.

\begin{figure}[h]
    \centering

        \includegraphics[width=\linewidth]{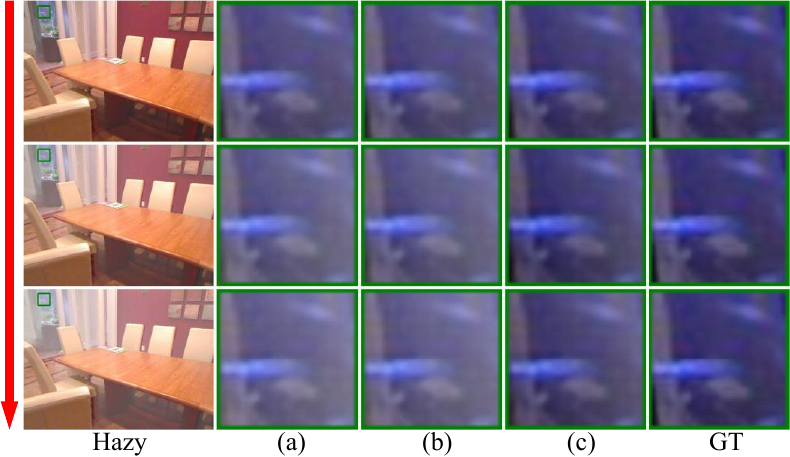}

        \caption{
            Visual comparisons across three different haze levels. (a) Baseline~\cite{fang2025guided} without depth; (b) Baseline enhanced by~\cite{ye2025prompthaze}; (c) Baseline enhanced by our method. All images are shown with consistent cropping for fair comparison. Compared to degraded hazy inputs, our method consistently restores clearer structures and colors, producing the closest match to ground truth.
        }
        
    \label{fig:vis-across-haze}
\end{figure}

{In this work, we demonstrate that incorporating depth features from large pretrained monocular depth estimation models facilitates more consistent RGB representations across varying haze intensities (Fig.~\ref{fig:tsne-vis},~\ref{fig:intro-multi-haze-compare},~\ref{fig:vis-across-haze}). Motivated by this insight, we propose a generalizable dehazing framework that hierarchically integrates depth and image features via a gated fusion mechanism. Our approach explicitly leverages the structural properties of depth representations and their interplay with appearance cues, yielding a principled and adaptive solution for haze removal. Extensive experiments further validate the effectiveness and strong generalization capability of our framework.}

In summary, our key contributions are threefold:

{1. We perform intuitive analysis to demonstrate that depth features extracted from large-scale depth estimation models effectively align RGB representations across varying haze levels, motivating their use to enhance image dehazing.}

{2. We propose a general framework that can be directly integrated into existing dehazing approaches, seamlessly combining depth and image features. Within this framework, we design an RGB-Depth Fusion Block to transfer reliable depth features from large pretrained depth estimation models.}

3. Extensive experiments on multiple datasets and scenarios verify the performance gains of our method, achieving state-of-the-art results.

\section{Related work}

\begin{figure*}[htbp]
    \centering
    \includegraphics[width=\linewidth]{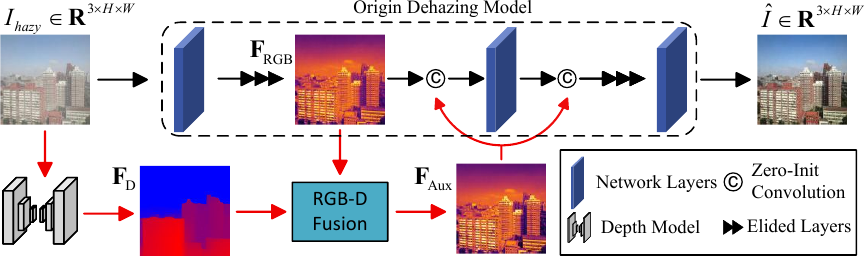} 
    \caption{Overview of the proposed framework. The black arrows illustrate the original data flow of the RGB model, while the red arrows indicate the additional flow introduced by our method. Specifically, we retain the original dehazing model and incorporate an RGB-Depth Fusion Module to fuse RGB and depth features. Further details of this module will be provided in the following sections. Zoom in for a better view.}

    \label{fig:model_arch}
\end{figure*}

\subsection{Image Dehazing}
Image dehazing methods based on physical models often rely on the atmospheric scattering model \cite{mccartney1976optics}:

\begin{equation}
    I(x) = J(x) t(x) + A (1 - t(x))
    \label{eq:asm}
\end{equation}
where $I(x)$ represents the observed hazy image, $J(x)$ denotes the underlying haze-free image to be recovered, $A$ is the global atmospheric light, and $t(x)$ is the transmission map that quantifies light attenuation due to scattering. For instance, \citeauthor{fattal2008single} cast image dehazing as an optimization problem, assuming that surface shading variations and transmission are locally uncorrelated. \citeauthor{he2010single} introduce the dark channel prior to facilitate an approximation of the transmission map. \citeauthor{zhu2015fast} propose a fast haze removal method based on a color attenuation prior, modeling the depth information using a linear model. With the rise of deep learning, particularly convolutional neural networks and transformers, researchers began leveraging deep neural networks for image dehazing. These methods learn complex features directly from data, overcoming the limitations of hand-crafted priors and achieving notable improvements in accuracy. By training on various datasets, deep learning-based approaches can generalize better to diverse scenes, leading to more effective haze removal compared to tradition methods~\cite{cai2016dehazenet,liu2019griddehazenet,chen2019gated,qin2020ffa,ren2018gated,dong2020physics,wu2021contrastive,chen2021psd,zheng2021ultra,tu2022maxim,guo2022image,qiu2023mb,chen2023dea,zheng2023curricular,zheng20234k,chen2024dea,lu2024mixdehazenet,cui2024omni}.


\subsection{Monocular Depth Estimation}

Depth estimation, a key multimodal task, predicts scene depth from a single RGB image and is commonly formulated as a regression problem. Early breakthroughs include coarse-to-fine networks~\cite{NIPS2014_7bccfde7} and encoder-decoder architectures~\cite{laina2016deeperdepthpredictionfully} that improved spatial detail. To enhance generalization, MiDaS~\cite{ranftl2020robustmonoculardepthestimation} was trained on diverse datasets, boosting robustness across scenes. With the rise of large-scale pretraining and generative models, the field has witnessed a new wave of innovation. \citeauthor{ke2023repurposing} proposed a method that utilizes a pretrained diffusion model and relies exclusively on synthetic data for training, demonstrating the surprising potential of generative priors in geometric understanding. More recently, \citeauthor{yang2024depthv2} introduced Depth Anything, which builds upon DINOv2~\cite{oquab2023dinov2} and self-supervised learning to achieve state-of-the-art zero-shot performance on diverse benchmarks. Alongside Depth Anything, several concurrent efforts have also aimed to push the boundaries of zero-shot monocular depth estimation, including techniques that leverage large-scale foundation models, synthetic data augmentation, and novel distillation strategies~\cite{he2025distill,bochkovskii2024depth,he2024lotus}.

\section{Methods}

In this section, we propose a generalizable dehazing framework that directly integrates into existing vision backbones and incorporates depth cues through modular and hierarchical fusion. Guided by a two-stage training strategy and zero-initialized convolution layers, the design ensures stable optimization while preserving the original RGB performance. At its core lies the RGB-Depth Fusion Module, which enables context-aware, geometry-guided enhancement via adaptive gating and cross-modal interaction, selectively injecting structural information into the RGB stream.

\subsection{Overall Framework} 

\begin{figure*}[h]
    \centering
    \includegraphics[width=\linewidth]{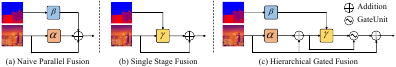}
    \caption{
        Comparison of RGB$\text{-}$Depth fusion strategies. 
        (a) Naive Parallel Fusion: RGB and depth features are processed independently.
        (b) Single Stage Fusion: The fusion of RGB and depth features is performed only at the final layer.
        (c) Hierarchical Gated Fusion (Ours): Features are enhanced through cross-layer interaction, gated modulation, and residual connections, allowing depth information to be selectively injected into the RGB stream.}
    \label{fig:hgdf}
\end{figure*}

Given a hazy image $I_{\text{hazy}} \in \mathbb{R}^{3 \times H \times W}$, our framework first processes it through a standard RGB-based dehazing model to extract baseline visual features, denoted as $\mathbf{F}_{\text{RGB}}$. In parallel, the input is forwarded to a pretrained monocular depth estimation network to obtain depth feature $\mathbf{F}_{\text{D}}$~(Fig.~\ref{fig:model_arch}). 

To incorporate geometric priors from the depth modality into the visual restoration process, we introduce an {RGB-Depth Fusion Module}, which adaptively merges the RGB features $\mathbf{F}_{\text{RGB}}$ with the depth features $\mathbf{F}_{\text{D}}$ to produce auxiliary features $\mathbf{F}_{\text{Aux}}$.  Inspired by the success of ControlNet~\cite{zhang2023adding}, we adopt zero-initialized convolutional layers to safely incorporate the auxiliary depth branch. Specifically, prior to each fusion operation, we insert a $1 \times 1$ convolutional layer whose weights are initialized to zero. This design choice effectively mitigates unstable training behavior and ensures that the initial performance of the pretrained RGB backbone is preserved.

Formally, given an input RGB feature map $\mathbf{F}_{\text{RGB}}$ and a depth-guided auxiliary feature map $\mathbf{F}_{\text{Aux}}$, the output feature $\mathbf{F}_{\text{RGB}}^{\prime}$ after the fusion block is computed as:
\begin{equation}
    \mathbf{F}_{\text{RGB}}^{\prime} = \phi\left( \mathbf{F}_{\text{RGB}} + \mathbf{W}_{\text{pre}} \cdot \mathbf{F}_{\text{Aux}} \right) + \mathbf{W}_{\text{post}} \cdot \mathbf{F}_{\text{Aux}},
\end{equation}
where $\phi(\cdot)$ denotes the original RGB processing block (e.g., a residual block), and $\mathbf{W}_{\text{pre}}$ and $\mathbf{W}_{\text{post}}$ are zero-initialized convolutional (ZC) layers. This residual-style formulation enables the network to gradually incorporate depth cues.

This modular fusion strategy enables the network to leverage complementary depth cues. The enhanced features are further propagated through the remaining layers of the RGB backbone, ultimately yielding the restored image $\hat{I} \in \mathbb{R}^{3 \times H \times W}$. Notably, our design maintains the architectural integrity of the original dehazing model, while augmenting it with an auxiliary depth stream that improves restoration performance with minimal overhead. Additionally, following~\cite{podell2023sdxl,zhang2023adding,mou2024t2i,hu2024animate} we adopt a multi-stage training strategy. In the first stage, we train the original RGB-based model until it achieves stable performance. Then, in the second stage, we activate the fusion modules and fine-tune the entire network using a cosine annealing with restarts schedule. This staged approach allows the network to retain its core capability while progressively incorporating depth information. This is where the zero-initialized convolution layers play a crucial role, as they prevent the depth features from disrupting the original dehazing process during the initial training phase. Experimental results also demonstrate the effectiveness of our design, showing consistent improvements across different architectures~(Tab.~\ref{table:rw2ah},~\ref{table:sots-in}).

\subsection{RGB-Depth Fusion Module} 
\label{sec:rbgd_gftf}

Effectively combining color and geometry is essential for advancing single image dehazing. While RGB images offer rich photometric details, depth maps provide structural priors that enhance edge sharpness and spatial coherence. However, existing designs often fall short in leveraging these complementary cues: two-stream architectures typically lack sufficient interaction, limiting the influence of geometric information, and single-stage fusion is inadequate for effectively learning cross-modal representations. 

To overcome this, we propose a Hierarchical Fusion Strategy that enables multi-level integration of modalities throughout the network(Fig.~\ref{fig:hgdf}).
By progressively injecting depth-informed structural cues into the RGB feature hierarchy, our approach enables the network to jointly exploit photometric and geometric information at multiple stages of representation learning. Central to this design is the Hierarchical Gated Depth Fusion (HGDF) module of which the detailed architecture will be elaborated in the next section. 

We also incorporate an Adaptive Spatial Gating (ASG) to further dynamically  suppressing unreliable or misaligned signals at the end of this block. The spatial gating map $\mathbf{M} \in [0,1]^{B \times 1 \times H \times W}$ is generated by applying a sigmoid activation to the output of the ASG function:

\begin{equation}
\mathbf{M} = \sigma \big( \mathrm{ASG}(\hat{\mathbf{F}}) \big),
\end{equation}
where $\mathrm{ASG}(\cdot)$ denotes a learnable mapping parameterized by conv-layers, $\hat{\mathbf{F}}$ denotes the output of the HGDF layers.

The final gated RGB feature is computed by spatially modulating the fused feature $\hat{\mathbf{F}}$ with the original RGB feature $\mathbf{F}_{{RGB}}$ as follows:

\begin{equation}
\hat{\mathbf{F}}_{\text{RGB}} = \hat{\mathbf{F}} \odot \mathbf{M} + \mathbf{F}_{\text{RGB}} \odot (1 - \mathbf{M}),
\end{equation}
where $\odot$ denotes element-wise multiplication with broadcasting along the channel dimension.

\begin{table*}[t]
    \centering
    \renewcommand{\arraystretch}{1.2}
    \scalebox{1}{
    \begin{tabular}{l|ccc|ccc|cc}
    \toprule
    \textbf{Methods} & PSNR$\uparrow$ & SSIM$\uparrow$ & PSNR-Y$\uparrow$ & LPIPS$\downarrow$ & FADE$\downarrow$ & NIQE$\downarrow$ & ($\Delta$)Params & ($\Delta$)FLOPS \\ 
    \midrule
 
    \midrule

    \textbf{MB-Taylorformer} & 19.49 & 0.5610 & 21.20 & 0.3808 & 0.8031 & 5.9032 & 2.66M & 24.36G \\
    \midrule 
    +SelfPromer & 19.67 & 0.5635 & 21.35 & 0.3783 & 0.8142 & 5.9310 & +703K & +9.24G \\
    +PromptHaze & \underline{19.83} & \underline{0.5666} & \underline{21.46} & \underline{0.3770} & \underline{0.7904} & \underline{5.4518} & +1.18M & +13.61G \\
    +Ours & \textbf{20.14} & \textbf{0.5726} & \textbf{21.75} & \textbf{0.3745} & \textbf{0.7765} & \textbf{5.4122} & +780K & +10.22G \\
    \midrule

    \midrule
    \textbf{DehazeFormer} & 20.36 & 0.5764 & 21.96 & 0.3721 & 0.7025 & \underline{5.7925} & 4.63M & 36.02G \\
    \midrule
    +SelfPromer & {20.55} & \underline{0.5765} & {22.15} & \underline{0.3673} & 0.7099 & 5.8888 & +703K & +8.34G \\
    +PromptHaze & \underline{20.68} & \textbf{0.5807} &\underline{22.33}  & 0.3766 &\underline{0.7015}  & 5.8202 & +1.18M & +12.44G \\
    +Ours & \textbf{20.92} & \textbf{0.5807} & \textbf{22.51} & \textbf{0.3604} & \textbf{0.6832} & \textbf{5.7583} & +780K & +10.22G \\
    \midrule

    \midrule
    \textbf{OKNet} & 21.73 & 0.6015 & 23.28 & 0.3437 & 0.4756 & 5.5214 & 4.42M & 30.28G \\
    \midrule
    +SelfPromer & 21.80 & 0.6056 & 23.33 & 0.3483 & 0.4622 & 5.4235 & +1.09M & +13.82G \\
    +PromptHaze & \underline{21.91} & \underline{0.6082} & \underline{23.42} & \underline{0.3409} & \underline{0.4613} & \underline{5.3346} & +1.91M & +21.33G \\
    +Ours & \textbf{22.12} & \textbf{0.6129} & \textbf{23.63} & \textbf{0.3387} & \textbf{0.4483} & \textbf{5.3091} & +1.22M & +15.45G \\
    \midrule
    
    \midrule
    \textbf{CoA} & 21.83 & \underline{0.6130} & 23.35 & 0.3571 & 0.4207 & 5.4235 & 3.75M & 3.06G \\
    \midrule
    +SelfPromer & 21.90 & 0.6127 & 23.42 & \underline{0.3531} & \underline{0.4202} & 5.3951 & +1.09M & +1.49G \\
    +PromptHaze & \underline{22.04} & 0.6124 & \underline{23.55} & 0.3574 & 0.4218 &\underline{5.3320} & +1.91M & +1.96G \\
    +Ours & \textbf{22.18} & \textbf{0.6155} & \textbf{23.73} & \textbf{0.3497} & \textbf{0.4082} & \textbf{5.3273} & +1.22M & +1.59G \\
    \midrule

    \midrule
    \textbf{SGDNet} & 22.18 & 0.6232 & 23.71 & {0.3522} & {0.3952} & 5.0451 & 10.979M & 40.54G \\
    \midrule 
    +SelfPromer & {22.32} & {0.6290} & {23.84} & 0.3487 & \underline{0.3937} & \underline{5.0681} & +1.09M & +13.82G \\
    +PromptHaze & \underline{22.39} & \underline{0.6330} & \underline{23.92} & \underline{0.3452} & 0.3959 & 5.0994 & +1.91M & +21.33G \\
    +Ours & \textbf{22.52} & \textbf{0.6390} & \textbf{24.05} & \textbf{0.3378} & \textbf{0.3879} & \textbf{5.0337} & +1.22M & +15.45G \\
    \bottomrule
    \end{tabular}
    }
    \caption{Quantitative evaluation on RW$^2$AH dataset. Metrics with $\downarrow$ indicate that lower values are better, while those with $\uparrow$ indicate higher is better. The best results are highlighted in bold, and the second-best are underlined. FLOPs are computed using an input resolution of \textbf{224$\times$224}. Our method consistently outperforms all previous methods across all metrics, demonstrating its effectiveness in enhancing dehazing performance.}
    \label{table:rw2ah}
\end{table*}

\subsection{Hierarchical Gated Depth Fusion Module}

To effectively fuse color and geometric cues, the proposed Hierarchical Gated Depth Fusion (HGDF) module hierarchically integrates RGB and depth features through four key components: a Color Feature Encoder (CFE) with self-attention to capture long-range dependencies within the RGB representation itself; a Depth-Aware Interaction (DAI) module using cross-attention to inject structural cues from depth; an Adaptive Channel Gating (ACG) to modulate depth contributions; and a Feed-Forward Network (FFN) for further refinement via channel-wise convolutions.

We now describe the process in detail. Given an RGB feature map $\mathbf{F}_{\text{RGB}} \in \mathbb{R}^{B \times C \times H \times W}$ and a depth feature map $\mathbf{F}_{\text{D}} \in \mathbb{R}^{B \times C \times H \times W}$, the module operates as follows:

\begin{figure*}[h]
    \centering
    \includegraphics[width=\linewidth]{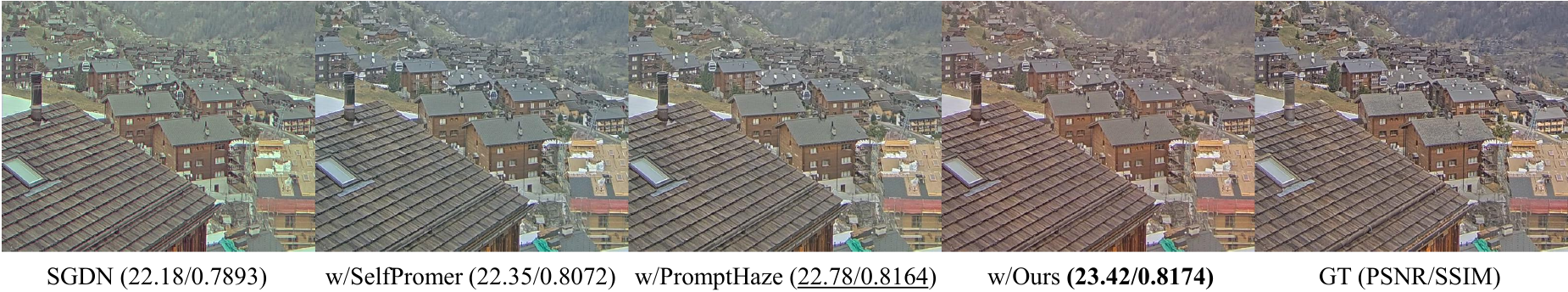} 
    \caption{Qualitative comparison of dehazing results, showing (from left to right) the input RGB image, baseline prediction, results enhanced by SelfPromer, PromptHaze, and our method, followed by the ground truth. Compared to prior approaches, our method better preserves color realism and produces more natural images. Additional visual results are included in the appendix.}
    \label{fig:exp_visual_rw2}
\end{figure*}

\paragraph{Color Feature Encoder.}
To enhance contextual understanding within the RGB modality, we apply a self-attention mechanism to the input RGB features. This allows the model to capture long-range dependencies and aggregate semantic information more effectively across the spatial domain. Formally, given the input RGB feature map $\mathbf{F}_{\text{RGB}}$, the attention-enhanced representation is computed as:

\begin{equation}
\mathbf{F}_{\text{RGB}}^{(A)} = \mathrm{MHSA}(\mathbf{F}_{\text{RGB}}),
\end{equation}
where $\mathrm{MHSA}(\cdot)$ denotes the multi-head self-attention operation, which computes pairwise interactions between spatial locations in the RGB feature map.

\paragraph{Depth-Aware Interaction.} To incorporate geometric priors into the RGB representation, we propose a channel-wise cross-attention mechanism, where the depth modality acts as a guidance signal to refine RGB features. Specifically, attention is computed along the channel axis, rather than the spatial domain, enabling semantically-aware recalibration of RGB features under geometric supervision.

Let $\mathbf{F}_{\text{RGB}}^{(A)}$ and $\mathbf{F}_{\text{D}}$ denote the attention-enhanced RGB features and depth features, respectively. We reshape both tensors to $\mathbb{R}^{B \times HW \times C}$ to facilitate attention computation along the channel dimension.

Following common practices in prior works, we adopt a two-layer convolutional embedding for the attention components, where each projection weight $\mathbf{W}^{Q}$, $\mathbf{W}^{K}$, and $\mathbf{W}^{V}$ consists of a $1 \times 1$ convolution followed by a $3\times3$ convolution:$\mathbf{Q}_{\text{RGB}}=\mathbf{F}_{\text{RGB}}^{(A)} \mathbf{W}_{1\times1}^{Q_1} \mathbf{W}_{3\times3}^{Q_2}$, $\mathbf{K}_{\text{D}} = \mathbf{F}_{\text{D}} \mathbf{W}_{1\times1}^{K_1} \mathbf{W}_{3\times3}^{K_2}$, $\mathbf{V}_{\text{RGB}} = \mathbf{F}_{\text{RGB}}^{(A)} \mathbf{W}_{1\times1}^{V_1} \mathbf{W}_{3\times3}^{V_2}$.

Then we calculate the channel-wise attention as:

\begin{equation}
    \mathbf{F}^{(X)}_{\text{RGB}} = \mathrm{Softmax}\left( \frac{\mathbf{Q}_{\text{RGB}} \mathbf{K}_{\text{D}}^\top}{\sqrt{d}} \right) \mathbf{V}_{\text{RGB}},
\end{equation}

Compared to~\cite{tsai2019multimodal}, which uses the auxiliary modality for both keys and values, our asymmetric design uses RGB as query and value, and depth as key, enabling task-aware refinement of RGB features with depth guidance while preserving semantic continuity.

\paragraph{Adaptive Channel Gating and Feed-Forward Network.}
Finally, we introduce a gated feature modulation strategy based on channel-wise attention. Specifically, a reshaped global context vector is extracted via global average pooling (GAP), followed by a channel-wise convolution and sigmoid activation to generate adaptive modulation weights:

\begin{equation}
\mathbf{g} = \sigma\left( \mathrm{MLP}\left( \mathrm{GAP} (\mathbf{F}^{(X)}_{\text{RGB}}) \right) \right),
\end{equation}

The gated features are then combined with the original RGB self-attention features through a residual fusion path:

\begin{equation}
\hat{\mathbf{X}} = \mathbf{F}_{\text{RGB}}^{(A)} + \mathbf{g} \odot \mathbf{F}^{(X)}_{\text{RGB}},
\end{equation}

To further refine the fused representation, we employ a lightweight dynamic feed-forward network (D-FFN) consisting of two $1 \times 1$ convolution layers interleaved with GELU activation:

\begin{equation}
\hat{\mathbf{F}} =  \mathbf{W}_{1\times1}^{(2)}\big( \mathrm{GELU}\big( \mathbf{W}_{1\times1}^{(1)}(\hat{\mathbf{X}}) \big) \big) + \hat{\mathbf{X}},
\end{equation}

The residual connection in the FFN promotes stable optimization and preserves the fused features~\cite{he2016deep}.

\begin{table}[htbp]
    \centering
    \renewcommand{\arraystretch}{1.14}
    \scalebox{1}{
    \begin{tabular}{l|ccc}
    \toprule
    \textbf{Methods} & PSNR & SSIM & PSNR-Y \\
    \hline\midrule
    \textbf{MB-Taylorformer} & 37.74 & 0.9901 & 39.30 \\ \midrule
    +SelfPromer & 38.48 & 0.9920 & 40.02 \\
    +PromptHaze & \underline{38.94} & \underline{0.9927} & \underline{40.48} \\
    +Ours & \textbf{39.33} & \textbf{0.9930} & \textbf{40.86} \\
    \midrule
        \midrule

    \textbf{DehazeFormer} & 38.14 & 0.9916 & 39.69 \\ \midrule
    +SelfPromer & 39.33 & 0.9923 & 40.90 \\
    +PromptHaze & \underline{39.71} &\underline{0.9928} & \underline{41.28} \\
    +Ours & \textbf{40.10} & \textbf{0.9935} & \textbf{41.63} \\
    \midrule
        \midrule

    \textbf{OKNet} & 37.54 & 0.9859 & 39.22 \\ \midrule 
    +SelfPromer & \underline{38.29} & \underline{0.9917} & 39.81 \\
    +PromptHaze & 38.22 & 0.9916 & \underline{39.84} \\
    +Ours & \textbf{38.70} & \textbf{0.9918} & \textbf{40.23} \\
    \midrule
    \midrule

    \textbf{CoA} & 35.85 & 0.9817 & 37.59 \\ \midrule 
    +SelfPromer & 36.39 & 0.9871 & 38.14 \\
    +PromptHaze & \underline{36.91} & \underline{0.9877} & \underline{38.62} \\
    +Ours & \textbf{37.77} & \textbf{0.9886} & \textbf{39.56} \\
    \midrule
        \midrule

    \textbf{SGDNet} & 38.97 & 0.9919 & 40.51 \\ \midrule 
    +SelfPromer & \underline{39.49} & \underline{0.9925} & \underline{41.10} \\
    +PromptHaze & 39.34 & 0.9919 & 40.92 \\
    +Ours & \textbf{39.99} & \textbf{0.9927} & \textbf{41.51} \\
    \bottomrule
    \end{tabular}
    }
    \caption{Quantitative evaluation on SOTS (Indoor).}
    \label{table:sots-in}
\end{table}

\section{Experiments}

\subsection{Experiment setup}

\subsubsection{Dataset}

We evaluate our method on both real-world and synthetic datasets~\cite{li2019benchmarking,fang2025guided}. The RW$^{2}$AH dataset comprises 1,406 training pairs and 352 test pairs. For synthetic data, we adopt the SOTS subsets: ITS (13,990 indoor image pairs), OTS (313,950 outdoor image pairs), and RESIDE-6K, which contains a mix of indoor and outdoor scenes, for comprehensive evaluation.

\subsubsection{Implementation Details}

We follow standard training protocols for image restoration. The initial learning rate is set according to each baseline. A cosine annealing with restarts scheduler is employed to decay the learning rate down to $1 \times 10^{-8}$, then reset and decay again in cycles. Training is conducted on 8 NVIDIA RTX 4090 GPUs, using gradient accumulation to support large batch sizes.

\subsubsection{Evaluation Metrics}

We evaluate the model performance using PSNR and SSIM, reported on both RGB (PSNR) and YCbCr (PSNR-Y) channels. For real-world datasets, we further employ LPIPS, FADE, and NIQE to assess perceptual quality. All methods are trained and tested under our unified settings to ensure fairness in comparison.

\begin{figure*}[h]
    \centering
    \includegraphics[width=\linewidth]{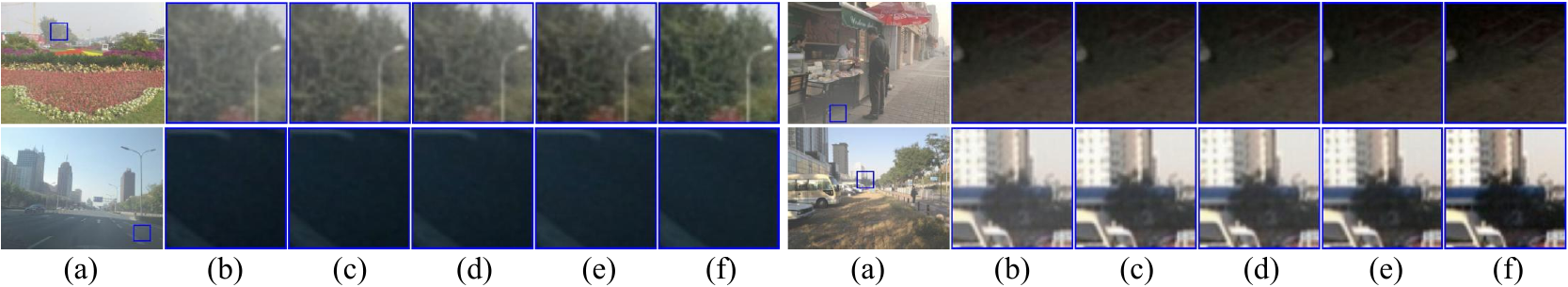}

    \caption{
        Visual comparison of haze removal results on the SOTS dataset.
        (a) The hazy input image; (b) The baseline prediction without enhancement; (c-e) Results enhanced by {SelfPromer}, {PromptHaze},
        and {our method}; (f) The ground truth.
        Our method not only recovers local details effectively but also achieves superior brightness restoration compared to other methods.
        }
    \label{fig:reside-compare}
\end{figure*}

\subsection{Experimental Results}

\begin{figure}[htbp]
    \centering
    \includegraphics[width=\linewidth]{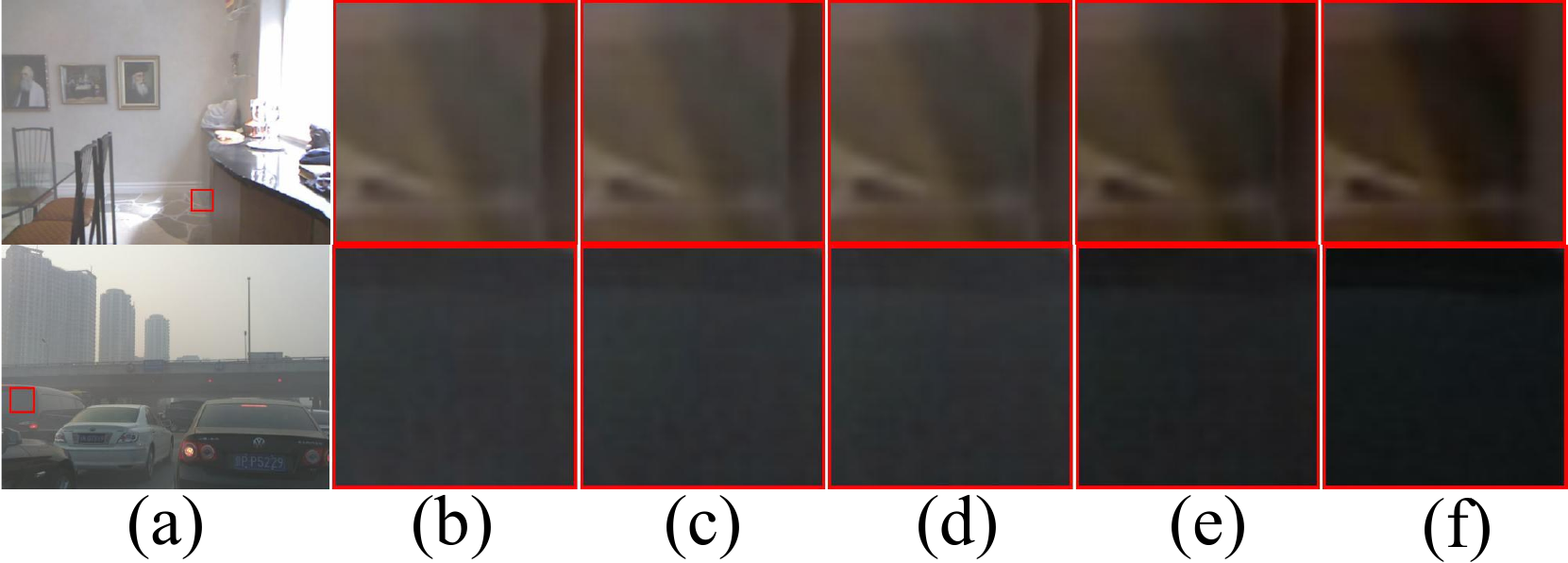}
    \caption{
        Visual comparison of haze removal results on the RESIDE-6K dataset. 
        (a) The hazy input image; (b) The baseline prediction without enhancement; (c-e) Results enhanced by {SelfPromer}, {PromptHaze}, 
        and {our method}; (f) The ground truth. Compared to previous methods, our approach produces improved chromatic consistency. 
    }
    \label{fig:reside-6k-compare}
\end{figure}

We evaluate our method on both synthetic and real-world dehazing benchmarks. Results in Tables~\ref{table:rw2ah} and~\ref{table:sots-in} show that our approach consistently outperforms existing methods, achieving the highest PSNR and SSIM across all datasets. Visual comparisons (Fig.~\ref{fig:exp_visual_rw2},~\ref{fig:reside-compare},~\ref{fig:reside-6k-compare}) further confirm that our method restores clearer images with more accurate details. Notably, our fusion strategy is the only one among evaluated variants that avoids performance degradation in all cases, demonstrating superior robustness and generalization~\cite{wang2024selfpromer,ye2025prompthaze}. We provide more quantitative and qualitative results in the appendix.

\subsection{Ablation study}

We conduct studies to evaluate the impact of key components. We use the SOTA method~\cite{fang2025guided} as baseline. We perform experiments on the RESIDE-6K dataset, which includes both indoor and outdoor scenes. Additional experiments  and discussion are provided in the appendix.

\subsubsection{Analysis of module in RGB-D Fusion Block} \quad We evaluate the effectiveness of each module in the RGB-D Fusion Block by systematically removing or modifying key components. The results are summarized in Table~\ref{tab:hgdf_module}.

\begin{table}[htbp]
    \centering
    \begin{adjustbox}{max width=0.9\textwidth}
    \begin{tabular}{@{}lcc@{}}
        \toprule
        {Module Combination} & {PSNR} & {SSIM} \\
        \midrule
        Baseline           & 29.12  & 0.9582 \\
        \midrule
        CFE + DAI                          & 29.48 & 0.9621 \\
        CFE + DAI + ASG                    &  \underline{29.59} & \underline{0.9624} \\
        CFE + DAI + ACG + FFN + ASG          & \textbf{29.79} & \textbf{0.9632} \\
        \midrule
        \bottomrule
    \end{tabular}
    \end{adjustbox}
    \caption{Quantitative evaluation of the impact of different components in RGB-D Fusion Block.}
    \label{tab:hgdf_module}
\end{table}

\subsubsection{Multi Stage Training} We further investigate the two-stage training strategy where the main dehazing backbone is frozen during the second stage. This approach still yields measurable performance improvements~(see Table~\ref{tab:finetune_choice}).

\begin{table}[h]
    \centering
    \begin{tabular}{lccc}
        \toprule
        Method & PSNR & SSIM  \\
        \midrule
        SGDNet & 29.12 & 0.9582 &  \\
        \midrule 
        +SelfPromer & 29.40 & {0.9625} \\
        +PromptHaze & \underline{29.54} & 0.9621 \\
        \midrule
        +Ours (Finetune Depth) & 29.53 & \underline{0.9626}  \\

        +Ours (Finetune all)& \textbf{29.79} & \textbf{0.9632}  \\
        \bottomrule
    \end{tabular}
    \caption{Quantitative comparison on the effect of fine-tuning the RGB branch during the second stage of training.    }
    \label{tab:finetune_choice}
\end{table}

\subsubsection{Performance on Deraining Task}\quad We extend our study to the deraining task, considering that rain, much like fog, obstructs light propagation and leads to image degradation~\cite{torralba2002depth}. We evaluate our approach on multiple public deraining benchmarks. Quantitative results  demonstrate the  effectiveness of our modeling (Tab.~\ref{tab:deraining_DRSformer}).

\begin{table}[htbp]
    \centering
    \begin{adjustbox}{max width=0.9\textwidth}
    \begin{tabular}{@{}cccccccccc@{}}
        \toprule
        \multirow{2}{*}{Method} 
        & \multicolumn{2}{c}{DID-Data}
        & \multicolumn{2}{c}{SPA-Data}
        
        \\
        
        \cmidrule(lr){2-3} 
        \cmidrule(lr){4-5} 
        \cmidrule(lr){6-7} 
        & PSNR-Y &  SSIM 
        & PSNR-Y &  SSIM 

        \\
        \midrule
        DRSformer 
        & 35.15 & 0.928
        & 48.92 & 0.990
        \\
        
        Ours    
        &  \textbf{35.40}   & \textbf{0.931}
        &  \textbf{49.40}   & \textbf{0.991}
        \\

        \midrule

        \bottomrule
    \end{tabular}
    \end{adjustbox}
    \caption{Comparison of performance on deraining task with SOTA methods~\cite{chen2023learning}.}
    \label{tab:deraining_DRSformer}
\end{table}

\subsection{Discussion}

We use depth information from large-scale pretrained models for dehazing without fine-tuning, reducing computational costs while maintaining strong performance. However, the large parameter size remains a limitation. A future direction could involve distilling a compact, efficient depth model for dehazing tasks. Further details are in the appendix.

\section{Conclusion}

In this work, we explore the consistency of depth information from large-scale pretrained models, which provide reliable priors for image dehazing. Building on this insight, we propose a flexible dehazing framework that seamlessly integrates depth cues with image features. By utilizing consistent depth information, our approach enhances the performance of existing SOTA models across diverse environments. Extensive experiments validate the effectiveness of our method, highlighting the potential of large-scale depth models to advance image dehazing and pave the way for future research in depth-assisted image restoration.

\subsubsection{Acknowledgments.}

We appreciate~\citeauthor{wang2024selfpromer,ye2025prompthaze} for generously providing their detail of implementation for comparison. Additionally, we are grateful to the authors of all referenced works for making their code and models publicly available, which facilitated our research and evaluation.

\bibliography{aaai2026}

\clearpage

\appendix

\twocolumn[
\begin{center}
    \LARGE\textbf{Supplementary Material: Can Large-Scale Depth Models Help With Image Dehazing?}
\end{center}
\vspace{2em}
]

\section{Outline of appendix}
This appendix presents both qualitative and quantitative analyses comparing depth feature representations derived from small-scale and large-scale models under hazy conditions. The investigation includes evaluations of feature activations and Kullback–Leibler (KL) divergence metrics. We systematically assess the influence of various depth estimation backbones, aiming to optimize the trade-off between dehazing accuracy and model complexity. An ablation study is conducted to demonstrate the efficacy of zero convolution modules in enhancing feature fusion and restoration performance. Experimental results on the RESIDE-6K benchmark substantiate the robustness of the proposed approach across diverse indoor and outdoor environments. Detailed descriptions of the training configurations, fusion strategies, and architectural components are provided to ensure reproducibility. Finally, we discuss the scalability constraints and deployment challenges associated with large pretrained depth models in real-world scenarios, and outline potential future research directions including model distillation and selective backbone freezing. These additional results and insights further demonstrate the effectiveness and adaptability of our proposed framework.

\begin{figure}[htbp] 
    \centering
    \begin{minipage}{0.49\linewidth}
        \centering
        \includegraphics[width=\linewidth]{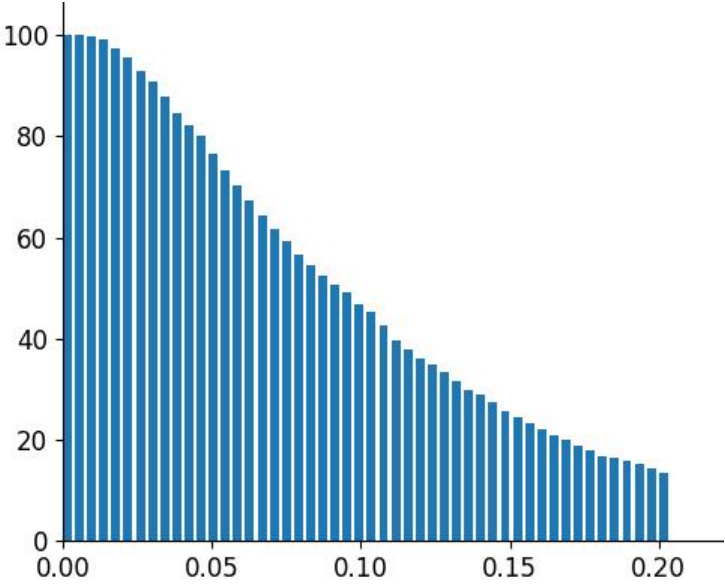}
    \end{minipage}
    \hfill
    \begin{minipage}{0.49\linewidth}
        \centering
        \includegraphics[width=\linewidth]{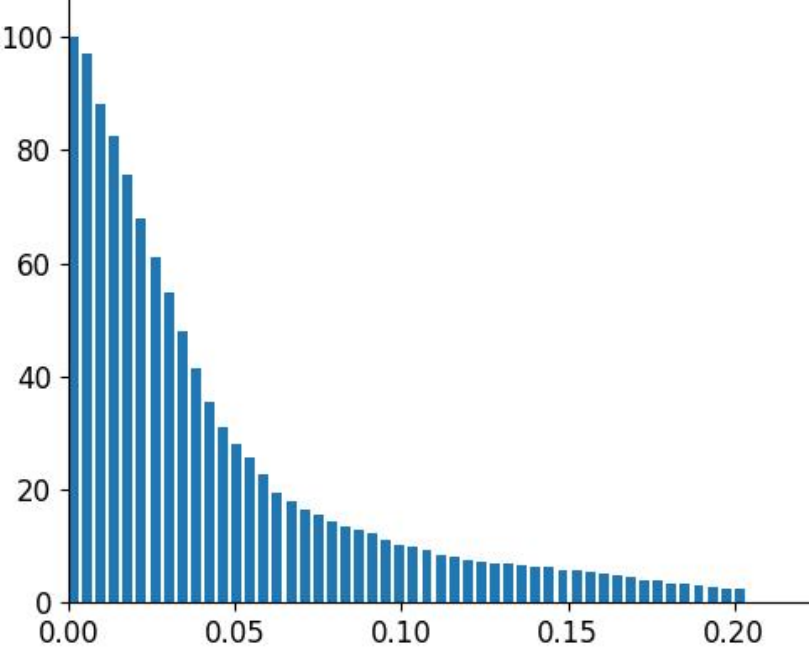}
    \end{minipage}
    \caption{KL divergence threshold analysis between the predicted depth of hazy images and the predicted depth of ground truth for two different depth models: \textbf{(a)} trained on limited data and \textbf{(b)} on large-scale data. The x-axis represents the KL divergence threshold, and the y-axis shows the proportion of values exceeding this threshold. The results indicate that the large-scale depth model, when applied to hazy images, produces depth predictions that are statistically closer to those derived from the ground truth.}
    \label{fig:kl-vis}
\end{figure}

\begin{figure*}[htbp]
    \centering
    \includegraphics[width=\linewidth]{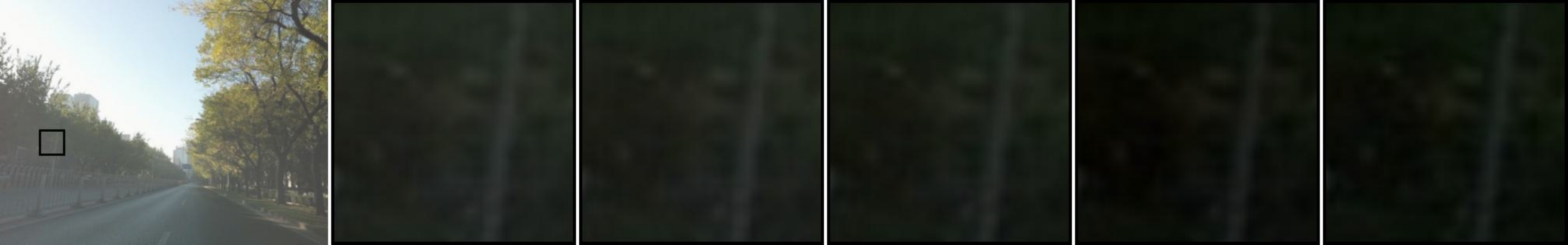}
    \includegraphics[width=\linewidth]{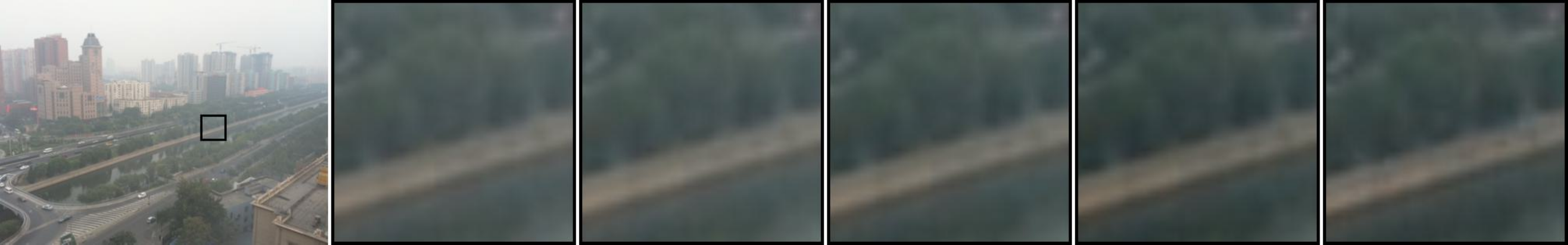}
    \includegraphics[width=\linewidth]{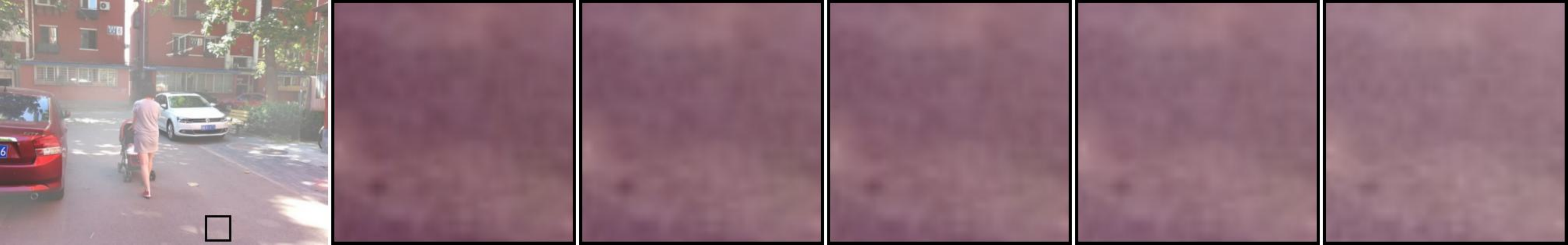}

    \caption{Indoor and outdoor scene. Comparison of results from SGDN, SelfPromer, PromptHaze, our method and GT.}

    \label{fig:outdoor}
\end{figure*}

\begin{table}[htbp]
    \centering
    \begin{adjustbox}{max width=\textwidth}
    \begin{tabular}{@{}cccccc@{}}
        \toprule
         Pre-ZC & Post-ZC & PSNR & SSIM   \\
        \midrule
        \multicolumn{2}{c}{Baseline} & 29.12 &0.9582\\
        \midrule
        \ding{55}&\ding{55}  & 29.22  & 0.9596   \\

         \checkmark&\ding{55} & \underline{29.64}  & \underline{0.9630 } \\
        \ding{55}& \checkmark   & 29.47  & 0.9622  \\
         \checkmark& \checkmark & \textbf{29.79} & \textbf{0.9632} \\

        \bottomrule
    \end{tabular}
    \end{adjustbox}
    \caption{Performance comparison of whether applying zero convolution at different stages. Cross mark means we directly use addition instead of zero convolution.}
    \label{tab:exp_zero_conv}
\end{table}

\begin{table}[h]
    \centering
    \renewcommand{\arraystretch}{1.2}
    \scalebox{1.0}{
    \begin{tabular}{l|c|c|c}
    \midrule
    \textbf{Backbone} & \textbf{LR} &\textbf{B.S.} & \textbf{Fusion Module}  \\
    \midrule
    MB-Taylor & $2 \times 10^{-4}$ & 32& MHCA   \\
    Dehaze-F & $3 \times 10^{-4}$ & 64 & DF-Block   \\
    MITNet & $2 \times 10^{-4}$ & 32& RAB  \\
    SADNet & $2 \times 10^{-4}$ &  64 &SDEB   \\
    OKNet & $4 \times 10^{-4}$ & 64&ResGroup   \\
    DAENet & $4 \times 10^{-4}$ &  16&DEBlock,DEBAlock   \\
    DNMGDT & $2 \times 10^{-4}$ & 64&Conv-L   \\
    CoA & $3 \times 10^{-4}$ &  64&ResidualBlock   \\
    SGDNet & $4 \times 10^{-4}$ & 16&HGDF  \\
    \midrule
    \end{tabular}
    }
    \caption{Training configurations for different backbones. Fusion Module indicates the modules we fuse. \textbf{B.S.} means batch size. All adopt the same two-stage training scheme.}
    \label{table:reside6k_setting}
\end{table}

\begin{table}[htbp]
    \centering
    \begin{adjustbox}{max width=\textwidth}
    \begin{tabular}{@{}cccccc@{}}
        \toprule
        Methods
         &  PSNR  & SSIM 
        & $\Delta$Param. 
        \\
        
        \midrule
        
        Baseline
        &  29.12 & 0.9582
        & -  \\
        
        \midrule
        w/ DAv2-S
        & 29.79  & 0.9632
        &  \multirow{2}{*}{25M} \\

        w/ DSv2-S
        & 29.69  & 0.9628
        &   \\
        \midrule
        w/ DAv2-M
        & 30.05  & 0.9650
        &  \multirow{2}{*}{98M}   \\
        w/ DSv2-M
        & 30.00  & 0.9646
        &   \\

        \midrule
         w/ DAv2-L
        & 30.19  & 0.9654
        &  \multirow{2}{*}{335M}  \\
        w/ DSv2-L
        & 30.27  & 0.9656
        &   \\
        \bottomrule
    \end{tabular}
    \end{adjustbox}
    \caption{Comparison of the performance of different backbones for the depth model with the baseline method. We report the additional  parameters of the depth model. The additional parameter of the fusion module is not included because of the relatively small size of the fusion module.}
    \label{tab:fusion_backbone_comparison}
\end{table}

\begin{figure}[h]
    \centering
    \includegraphics[width=0.95\linewidth]{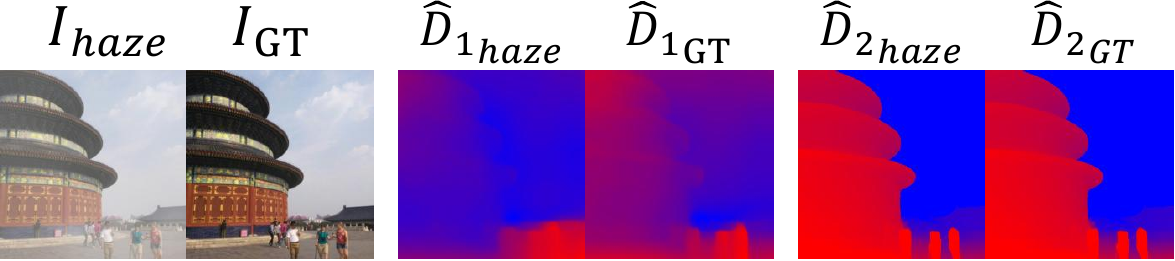}
    \caption{Depth maps predicted by two depth estimation models from hazy images and their corresponding ground truth images: a small-scale model ($\hat{D}_1$)~\cite{he2022ra} and a large-scale model ($\hat{D}_2$)~\cite{yang2024depthv2}.}
    \label{fig:depth_predictions}
\end{figure}

\begin{figure}[h] 
    \centering
    \includegraphics[width=0.95\linewidth,keepaspectratio]{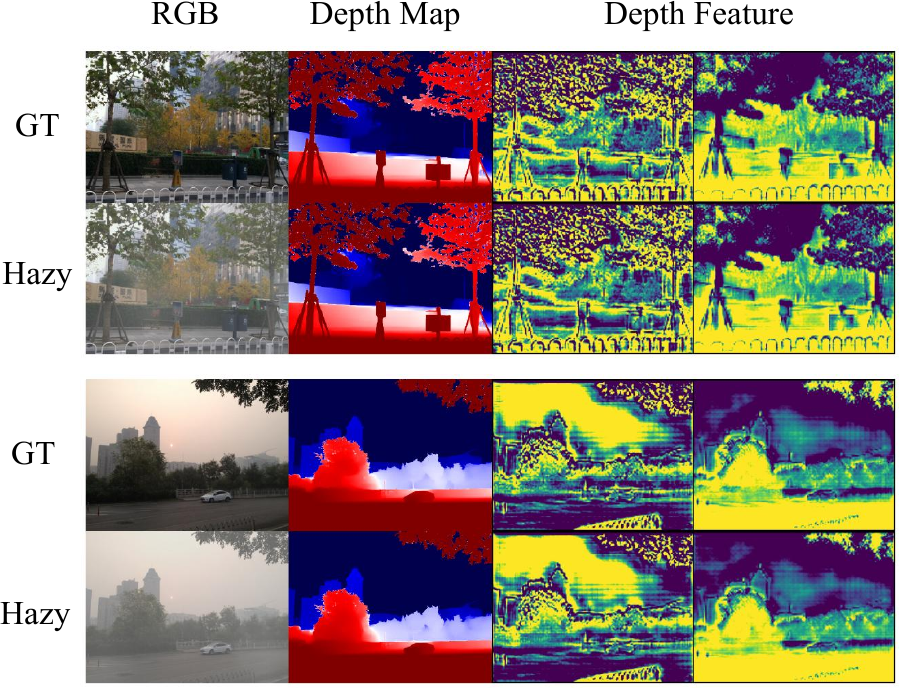} \\ 
    \caption{Depth map and feature visualization of \cite{yang2024depthv2} for the hazy image and ground truth image. The left side shows the ground truth and hazy image, the  column on the right displays its predicted depth map. The rightmost two columns show feature maps generated by intermediate layers from the corresponding image.}
    \label{fig:feature_vis}
\end{figure}

\section{Visualization of depth feature}

To further motivate our study, this section provides additional evidence supporting the robustness of large-scale pretrained depth models through both quantitative and qualitative comparisons. Existing depth-assisted dehazing methods~\cite{zhang2024depth,yang2022depth} typically rely on lightweight depth estimation networks to provide depth information. However, these models are often trained on small-scale datasets, resulting in limited generalization capability and unreliable depth predictions—particularly in challenging regions affected by noise, occlusions, and missing depth cues (see Fig.~\ref{fig:depth_predictions}). Consequently, the extracted depth features tend to be inconsistent and less effective in guiding the dehazing process, which ultimately restricts the overall performance of these methods.

To quantitatively validate this advantage, we also conduct a comparative analysis of the Kullback-Leibler (KL) divergence between depth predictions generated by small-scale versus large-scale models for both hazy images and their corresponding ground truth counterparts (Fig.~\ref{fig:kl-vis}). Furthermore, we present visualizations of depth maps and intermediate feature representations generated by the large-scale model for both hazy and ground-truth image (Fig.~\ref{fig:feature_vis}).

\begin{table}[htbp]
    \centering
    \renewcommand{\arraystretch}{1.2}
    \scalebox{1}{
    \begin{tabular}{l|ccc}
    \toprule
    \textbf{Methods} & PSNR & SSIM & PSNR-Y \\
    \midrule\midrule
    \textbf{MB-Taylorformer} & 31.32 & 0.9571 & 33.03 \\ \midrule
    +SelfPromer & \underline{31.40} & \textbf{0.9601} & \underline{33.14} \\
    +PromptHaze & 31.37 & \underline{0.9598} & 33.07 \\
    +Ours & \textbf{31.73} & \textbf{0.9601} & \textbf{33.45} \\
    \midrule    \midrule

    \textbf{DehazeFormer} & 32.24 & 0.9656 & 34.06 \\ \midrule
    +SelfPromer & \underline{32.39} & \underline{0.9665} & \underline{34.22} \\
    +PromptHaze & 32.24 & 0.9660 & 34.07 \\
    +Ours & \textbf{32.70} & \textbf{0.9666} & \textbf{34.58} \\
    \midrule    \midrule

    \textbf{OKNet} & 32.28 & 0.9664 & 34.03 \\ \midrule 
    +SelfPromer & 32.36 & 0.9617 & 34.08 \\
    +PromptHaze & \underline{32.49} & \underline{0.9664} & \underline{34.28} \\
    +Ours & \textbf{32.64} & \textbf{0.9666} & \textbf{34.43} \\
    \midrule    \midrule

    \textbf{CoA} & 31.76 & 0.9600 & 33.53 \\ \midrule 
    +SelfPromer & 32.00 & \underline{0.9614} & 33.76 \\
    +PromptHaze &\underline{ 32.04} & 0.9613 & \underline{33.77} \\
    +Ours & \textbf{32.30} & \textbf{0.9628} & \textbf{34.09} \\
    \midrule    \midrule

    \textbf{SGDNet} & \underline{32.39} & \underline{0.9603} & \underline{34.22} \\ \midrule 
    +SelfPromer & 32.19 & 0.9595 & 34.00 \\
    +PromptHaze & \underline{32.39} & 0.9597 & 34.21 \\
    +Ours & \textbf{32.70} & \textbf{0.9606} & \textbf{34.58} \\
    \bottomrule
    \end{tabular}
    }
    \caption{Quantitative evaluation on SOTS (Outdoor).}
    \label{table:sots-out}
\end{table}

\section{Training Implementation Details}

This section provides additional implementation details for our method, including training configurations and fusion strategies. We use the RESIDE6K dataset for training and evaluation, which contains a diverse set of hazy images with corresponding ground truth.

\subsection{Training and fusion configurations}

As shown in Table~\ref{table:reside6k_setting}, we summarize the experimental settings and the integration points of depth features. Generally, dehazing—being a low-level vision task—is composed of feature extraction and encoder-decoder structures with downsampling and upsampling layers. In our design, the depth features are fused specifically at the feature extraction layer to better guide the restoration process.

\begin{table}[htbp]
    \centering
    \renewcommand{\arraystretch}{1.2}
    \scalebox{1.1}{
    \begin{tabular}{l|ccc}
    \midrule
    \textbf{Method} & PSNR$\uparrow$ & SSIM$\uparrow$ & PSNR-Y$\uparrow$  \\ 
    \midrule\midrule
    \textbf{MB-Taylor} & 28.71 & 0.9606 & 30.29 \\ 
    \midrule 
    +SelfPromer & 28.89 & 0.9618 & 30.48   \\
    +PromptHaze & \underline{29.01} & \underline{0.9624} & \underline{30.60} \\
    +Ours & \textbf{29.34} & \textbf{0.9636} & \textbf{30.93}  \\
    \midrule
    \midrule

    \textbf{MITNet} & 28.80 & 0.9562 & 30.37 \\ \midrule 

    +SelfPromer & 29.03 & \underline{0.9597} & 30.60   \\
    +PromptHaze & \underline{29.35} & 0.9587  & \underline{30.94} \\
    +Ours & \textbf{29.61} & \textbf{0.9608} & \textbf{31.20} \\
    \midrule
    \midrule

    \textbf{OKNet} & 29.14 &  0.9642 & 30.71 \\
    \midrule

    +SelfPromer & 29.20 & \underline{0.9644} & 30.78 \\
    +PromptHaze & \underline{29.34} & \textbf{0.9646} & \underline{30.93} \\
    +Ours & \textbf{29.69} & \textbf{0.9646} & \textbf{31.28} \\
    \midrule
    \midrule

    \textbf{DAENet} & 29.06 & 0.9614 & 30.66 \\
    \midrule 
    +SelfPromer & 29.18 & \underline{0.9629} & 30.77 \\
    +PromptHaze & \underline{29.32} & 0.9627 & \underline{30.90} \\
    +Ours & \textbf{29.61} & \textbf{0.9635} & \textbf{31.20} \\
    \midrule
    \midrule

    \textbf{SADNet} & 28.58 & 0.9544 & 30.10 \\
    \midrule 
    +SelfPromer & 28.79 & 0.9556 & 30.38 \\
    +PromptHaze &\underline{28.93} & \underline{0.9568} & \underline{30.52} \\
    +Ours & \textbf{29.17} & \textbf{0.9580} & \textbf{30.76} \\
    \midrule
    \midrule

    \textbf{DNMGDT} & 27.81 & 0.9509 & 29.36 \\ 
    \midrule 
    \midrule

    +SelfPromer & 28.10 & 0.9516 & 29.67 \\
    +PromptHaze & \underline{28.29} & \underline{0.9526} & \underline{29.87} \\
    +Ours & \textbf{28.47} & \textbf{0.9531} & \textbf{30.03} \\
    \midrule
    \midrule

    \textbf{SGDNet} & 29.12 &  0.9582 & 30.70 \\
    \midrule 
    +SelfPromer & 29.40 & \underline{0.9625} & 30.99 \\
    +PromptHaze & \underline{29.54} & 0.9621 & \underline{31.13} \\
    +Ours & \textbf{29.79} & \textbf{0.9632} & \textbf{31.38} \\
    \midrule

    \end{tabular}
    }
    \caption{Performance and model complexity on the RESIDE-6K dataset. $\uparrow$ indicates that higher values are better, while $\downarrow$ indicates that lower values are better.}
    \label{table:reside6k_only}
\end{table}

\subsection{Analysis of Depth Model Backbone}
 We evaluate the impact of different depth estimation model backbones and their influence on dehazing performance (Tab.~\ref{tab:fusion_backbone_comparison}). Our results indicate that increasing the backbone capacity from Small (S) to Large (L) leads to improvements in both PSNR and SSIM, albeit at the cost of higher memory consumption. This suggests that model efficiency, rather than sheer size, may be more critical for practical deployment, especially in resource-constrained settings.

\subsection{Effectiveness of zero convolution.} We study the impact of zero convolution at different stages of our model, with results summarized in Table~\ref{tab:exp_zero_conv}. The experiments show that introducing zero convolution either before or after the fusion process improves performance over the baseline. In particular, applying zero convolution before fusion yields a clear gain, reaching a PSNR of 29.64~dB and SSIM of 0.9630. When applied at both stages, the performance is further improved to 29.79~dB and 0.9632, demonstrating the complementary benefits of zero convolution in both positions.

\subsection{More experiments on mixed scenarios} We evaluate our method on the RESIDE6K dataset, which contains both indoor and outdoor scenes. The results are shown in Table~\ref{table:reside6k_only}. Our method consistently outperforms existing methods across all datasets, achieving the highest performance. We also provide visual comparison to further demonstrate the effectiveness of our methods (See Fig.~\ref{fig:outdoor}). We provide the quantitative results of outdoor scene in Table~\ref{table:sots-out}. Our method achieves the best performance on the SOTS dataset, outperforming other methods by a significant margin.

\begin{figure*}[htbp] 
    \centering
    \includegraphics[width=0.95\linewidth,keepaspectratio]{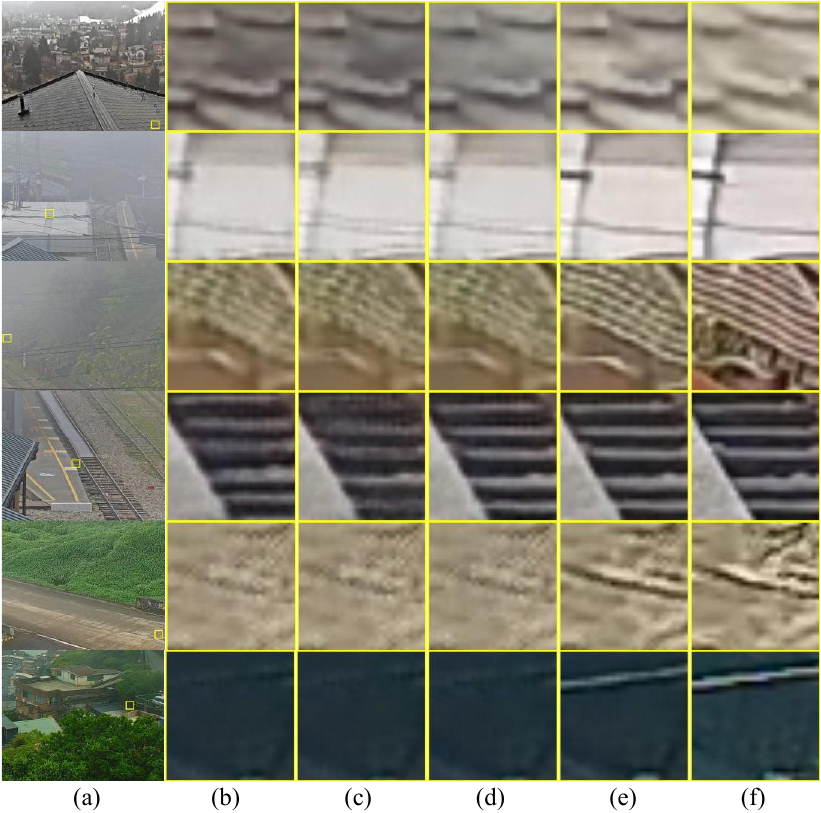}

    \caption{Visual comparison of results on RW2AH dataset. From left to right: Hazy image, SGDN, SelfPromer, PromptHaze, our method, and Ground Truth (GT). (a) The hazy input image; (b) The baseline prediction without enhancement; (c-e) Results enhanced by {SelfPromer}, {PromptHaze},
    and {our method}; (f) The ground truth.}
    \label{fig:rw2ah}
\end{figure*}

\subsection{More Discussion} 

Although we leverage large pretrained depth models without fine-tuning, their substantial size remains a barrier to deployment in resource-constrained environments. A viable future direction is to distill compact depth models that preserve critical structural cues for dehazing (See Table~\ref{tab:fusion_backbone_comparison}). Such models could benefit from knowledge distillation techniques that transfer representational power from large teacher networks to lightweight student models, enabling real-time performance on edge devices. Furthermore, integrating task-specific inductive biases—such as depth-gradient awareness or haze-density priors—into the student architecture may further enhance robustness without significantly increasing model complexity. Exploring architecture search or quantization-aware training could also complement the distillation process, paving the way for efficient and accurate depth-guided dehazing under practical constraints.

\end{document}